\documentclass[conference]{ieeetran}
\IEEEoverridecommandlockouts
\usepackage{cite}
\usepackage{amsmath,amssymb,amsfonts}
\usepackage{algorithmic}
\usepackage{graphicx}
\usepackage{siunitx}
\usepackage{bm}
\usepackage[colorlinks=true, linkcolor=blue, urlcolor=black, citecolor=blue]{hyperref}
\usepackage{multirow}
\usepackage{textcomp}
\usepackage{xcolor}
\usepackage{booktabs}
\usepackage[all]{hypcap}
\def\BibTeX{{\rm B\kern-.05em{\sc i\kern-.025em b}\kern-.08em
    T\kern-.1667em\lower.7ex\hbox{E}\kern-.125emX}}
\begin{document}

\title{Design, Modeling and Direction Control of\\ a Wire-Driven Robotic Fish Based on \\a 2-DoF Crank-Slider Mechanism\\
\author{Yita~Wang$^{1}$,~Chen~Chen$^{1}$,~Yicheng~Chen$^{1}$,~Jinjie~Li$^{1}$,~Yuichi~Motegi$^{2}$,~Kenji~Ohkuma$^{2}$,\\~Toshihiro~Maki$^{2}$,~Moju~Zhao$^{1}$}%
\thanks{$^{1}$Department of Mechanical Engineering, The University of Tokyo, Tokyo, 113-8656, Japan {\tt\small \{yita-wang, chen-chen, yicheng-chen, jinjie-li, chou\}@dragon.t.u-tokyo.ac.jp}}%
\thanks{$^{2}$Institute of Industrial Science, The University of Tokyo, Tokyo, 153-8505, Japan {\tt\small \{motegi, ohkuma, maki\}@iis.u-tokyo.ac.jp}}%
}

\maketitle
 
\begin{abstract}
Robotic fish have attracted growing attention in recent years owing to their biomimetic design and potential applications in environmental monitoring and biological surveys. Among robotic fish employing the Body--Caudal Fin (BCF) locomotion pattern, motor-driven actuation is widely adopted. Some approaches utilize multiple servo motors to achieve precise body curvature control, while others employ a brushless motor to drive the tail via wire or rod, enabling higher oscillation and swimming speeds. However, the former approaches typically result in limited swimming speed, whereas the latter suffer from poor maneuverability, with few capable of smooth turning. To address this trade-off, we develop a wire-driven robotic fish equipped with a 2-degree-of-freedom (DoF) crank–slider mechanism that decouples propulsion from steering, enabling both high swimming speed and agile maneuvering. In this paper, we first present the design of the robotic fish, including the elastic skeleton, waterproof structure, and the actuation mechanism that realizes the decoupling. We then establish the actuation modeling and body dynamics to analyze the locomotion behavior. Furthermore, we propose a combined feedforward–feedback control strategy to achieve independent regulation of propulsion and steering. Finally, we validate the feasibility of the design, modeling, and control through a series of prototype experiments, demonstrating swimming, turning, and directional control.

\end{abstract}

\section{Introduction}
\begin{figure}
    \centering
    \includegraphics[width=0.95\linewidth]{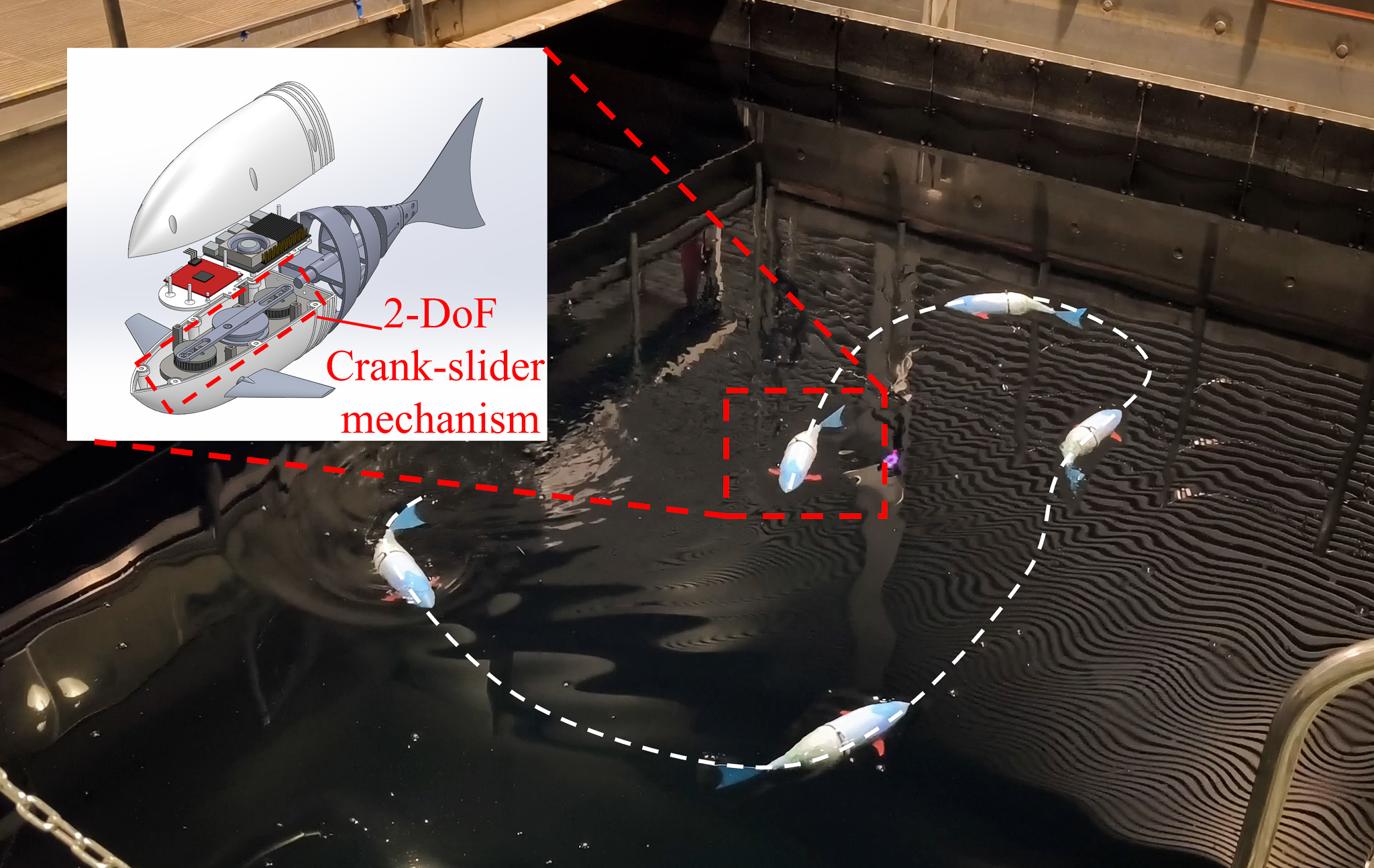}
    \caption{The biomimetic fish robot based on a 2-DoF crank-slider mechanism}
    \label{fig:first_fig}
    \vspace{-5mm}
\end{figure}

As global warming intensifies, ocean surveys—particularly biological ones—are attracting increasing attention. In current marine research, underwater robots such as autonomous underwater vehicles (AUVs) and remotely operated vehicles (ROVs) are widely employed for environmental monitoring, deep-sea exploration, and underwater manipulation \cite{underwater_manipulate_review, underwater_monitoring_review}, demonstrating their strong potential for ocean studies. However, propeller-driven robots like AUVs and ROVs generate significant noise and may threaten marine organisms and ecosystems, limiting their suitability for close-range observation of marine animals \cite{intro_noise_2020}. Biomimetic robots, with animal-like appearances and locomotion, present a promising alternative that may reduce disturbance to wildlife and the environment \cite{Wang2023Jellyfish, Chu2012_fish_review}. Among these designs, robotic fish offer advantages in actuation efficiency and potential for field applications \cite{MIT}. Despite progress in state-of-the-art designs, no existing system achieves both high swimming speed and precise maneuverability due to the properties of the actuation mechanism. To address this gap, we propose a crank–slider-based actuation mechanism for a wire-driven robotic fish through the design, modeling, and control of the mechanism and skeletal structure, and we validate the feasibility of this approach with a prototype, as illustrated in Fig.~\ref{fig:first_fig}.

Many studies have investigated the Body-Caudal Fin (BCF) propulsion method using different actuation strategies \cite{fish_review1, Chu2012_fish_review}. A common approach is the servo-driven multi-link body \cite{sailfish, multi_sevo_1, multi_servo_2_2012}, while other studies employ novel actuators such as McKibben artificial muscles, shape memory alloys (SMA), and hydraulic systems \cite{2Dof_soft_hydor_act, Gu2024, MIT}. These methods prioritize generating precise body curvature and large oscillation amplitudes, which enhance maneuverability. However, actuator limitations constrain oscillation frequency, resulting in relatively low tail-beat frequencies and making it difficult to achieve high-speed swimming. Another approach is to use brushless motors to actuate the tail, enabling higher oscillation frequencies \cite{orange_fish, tuna_bot, motor_act_slot, double-crank-slider}. Yet, most designs rely on mechanisms that convert continuous rotation into symmetric body oscillations, which restricts turning agility. Both strategies therefore face inherent limitations imposed by their actuation hardware. Therefore, in this work, we capitalize on the motor-based actuation, and expand the crank-slider mechanism into 2-DoF, which is able to decouple propulsion and steering, and thus able to gain both swimming speed and the maneuvering ability.

Modeling is essential for analyzing the actuation and dynamic behavior of robotic fish under different locomotion states, particularly in assessing how body motion affects hydrodynamic forces on position and attitude. For direct-drive multi-joint or caudal fin designs, modeling generally follows the conventional framework of multi-joint robotic systems \cite{general_model, motor_act_slot}. In contrast, soft actuators or elastic-body designs are often approximated by segmenting the continuously deforming body into multiple joints, which allows kinematic and dynamic analysis \cite{2Dof_soft_hydor_act}. In this work, we adopt a similar approximation for the elastic body of our robotic fish, with the goal of investigating the relationship among swimming frequency, oscillation amplitude, and the load distribution of the two motors in the proposed 2-DoF crank-slider mechanism.

Control is another key factor in enabling autonomous swimming and practical applications. Recent studies on robotic fish heading control have focused mainly on locomotion control, including orientation and trajectory tracking \cite{fish_review1, Sun2022FishRobots_review}. Single-actuator robots typically rely on preset signals to generate body undulation, while multi-DoF designs often employ central pattern generator (CPG) approaches \cite{MIT, 2Dof_soft_hydor_act}. However, locomotion control alone is insufficient for field applications due to limited tracking accuracy. To address this, perception systems such as artificial lateral line (ALL) sensors, inertial measurement units (IMU), and vision-based methods have been incorporated for improved orientation and trajectory control \cite{xieBoxfishCPG, Li2024, Yu2014VisionFish}. In this work, we first implement feedforward control to decouple propulsion and steering using the 2-DoF crank–slider mechanism, and then apply feedback control to achieve precise yaw direction following.

The main contributions of this work are summarized as follows:
\begin{itemize}
\item A novel actuation mechanism based on a 2-DoF crank–slider design that decouples propulsion and steering, along with the structural and waterproof design of a wire-driven robotic fish;
\item A modeling method for the proposed actuation mechanism and elastic body, incorporating fluid dynamics to estimate motor loads under different locomotion states, which also aids motor selection for prototypes;
\item A feedforward actuation control method to decouple propulsion and steering, followed by a feedback yaw direction control method;
\item Experimental validation of the proposed design, modeling, and control methods using a prototype, demonstrating swimming, turning, and direction control capabilities.
\end{itemize}

The remainder of this paper is organized as follows. Section \ref{sec:hardware} presents the hardware design. Section \ref{sec:modeling} describes the modeling of the actuation mechanism and body. Section \ref{sec:control} discusses feedforward actuation control and feedback direction control. Section \ref{sec:exp} reports the experimental results. Finally, Section \ref{sec:conclusion} concludes the paper.

\section{Hardware Design} \label{sec:hardware}

\subsection{Body Skeleton Design}
\begin{figure}[t]
    \centering
    \includegraphics[width=0.9\linewidth]{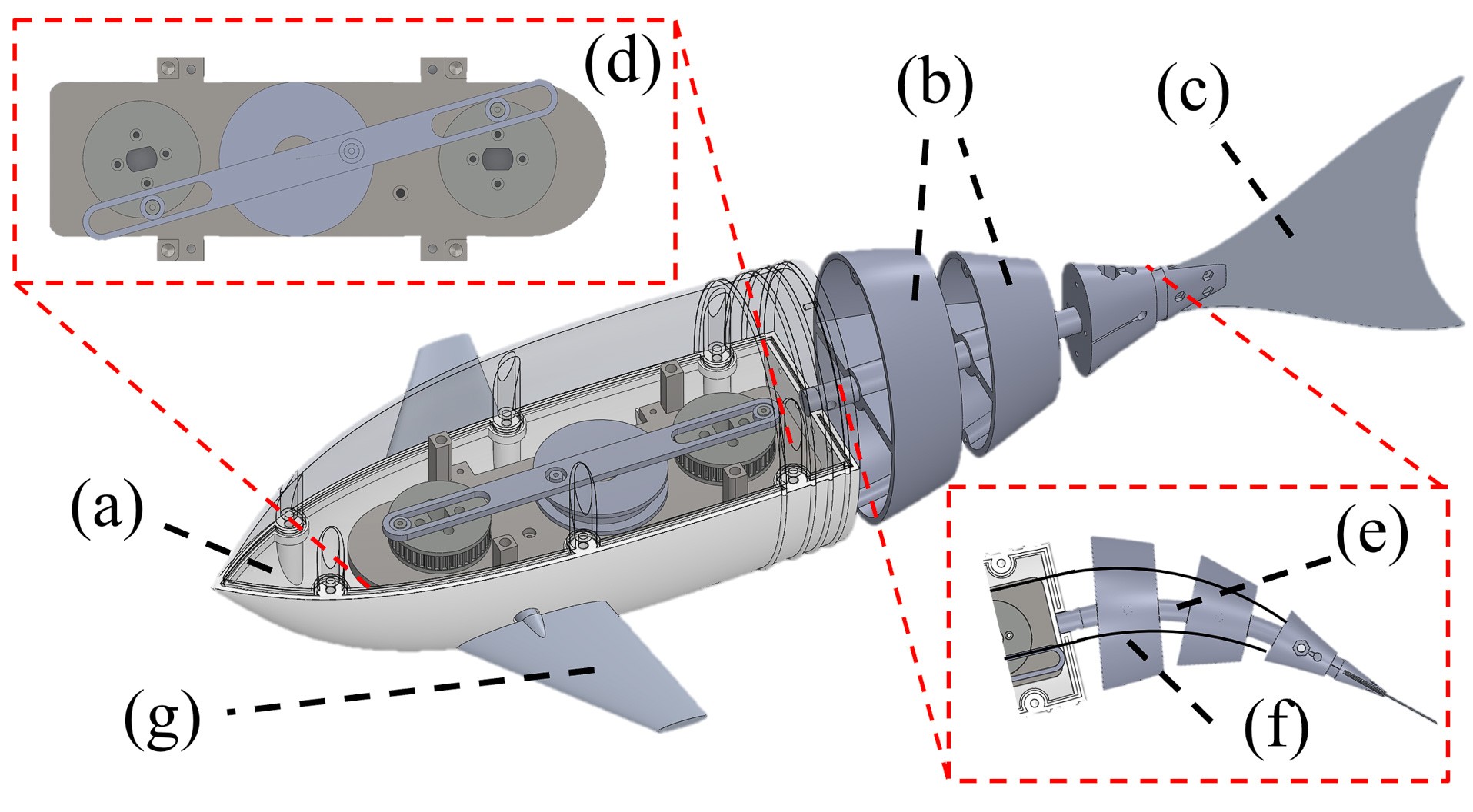}
    \caption{Hardware design of the robotic fish: (a) rigid head shell; (b) segmented body sections; (c) passive caudal fin; (d) 2-DoF crank–slider actuation mechanism; (e) elastic spine; (f) driving wire; (g) pectoral fins.}
    \label{fig:Mech_body}
    \vspace{-5mm}
\end{figure}

We adopt the \textit{subcarangiform} design of the BCF pattern, in which the rigid head comprises approximately half the body length, balancing speed and flexibility \cite{fish_review1}, as shown in Fig.~\ref{fig:Mech_body}. The skeleton consists of a rigid head and an elastic body formed by an elastic spine and three streamlined sections to reduce drag. Unlike conventional free joints, the elastic spine connects the sections, providing passive curvature control during bending. A driving wire fixed to the tail section actuates body oscillation. A pair of fixed pectoral fins enhances roll stability, while a compliant caudal fin improves actuation efficiency\cite{caudal_fin_soft}. In addition, we calculated the center of buoyancy (CB) and utilized ballast to adjust the weight distribution and the center of gravity (CoG), which further ensures the stability of the pitch and roll direction of the robot.

\subsection{Actuation Mechanism Design}

Recent wire-driven robotic fish typically use a single motor to drive a reel, simplifying the design. Two actuation schemes are common. In the first, a servo motor directly controls the reel rotation angle, which permits explicit control of oscillation amplitude and frequency; however, repeated acceleration and deceleration at each reversal incur substantial energy loss at higher frequencies. To mitigate this, many studies adopt a continuously rotating motor and a transmission that converts rotation into reciprocating wire motion. This approach supports higher oscillation frequencies but provides limited authority over amplitude and swimming direction. In summary, single-motor architectures tightly couple tail bias (mean angle) and oscillation, restricting maneuverability.

\begin{figure}[b]
\vspace{-5mm}
    \centering
    \includegraphics[width=\linewidth]{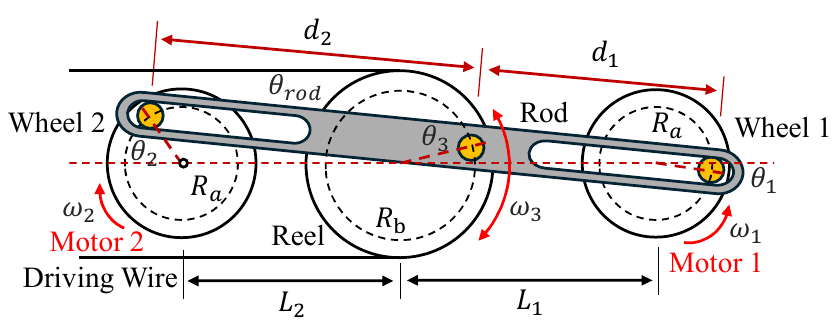}
    \caption{The 2-DoF crank-slider based actuation mechanism.}
    \label{fig:kine_act}
\end{figure}

To overcome these limitations, we propose a novel actuation mechanism based on extended crank-slider that decouples propulsion and steering, enabling turning at relatively high oscillation frequencies, as shown in Fig. \ref{fig:kine_act}. The mechanism comprises two driving wheels, each powered by an independent motor; their rotations are transmitted to the reel through a crank–slider linkage. The rod is joined to the reel, and the slots in the rod are slidably engaged with the wheels, forming a 2-DoF system. One motor sets the mean oscillation angle, while the other governs oscillation frequency via its rotation speed, achieving decoupled control. The modeling and control of this 2-DoF actuation are presented in Sections \ref{subsec:kine_act_mech} and \ref{subsec:feedforward}, respectively.

\subsection{Waterproof Design}
\begin{figure}[t]
    \centering
    \includegraphics[width=0.95\linewidth]{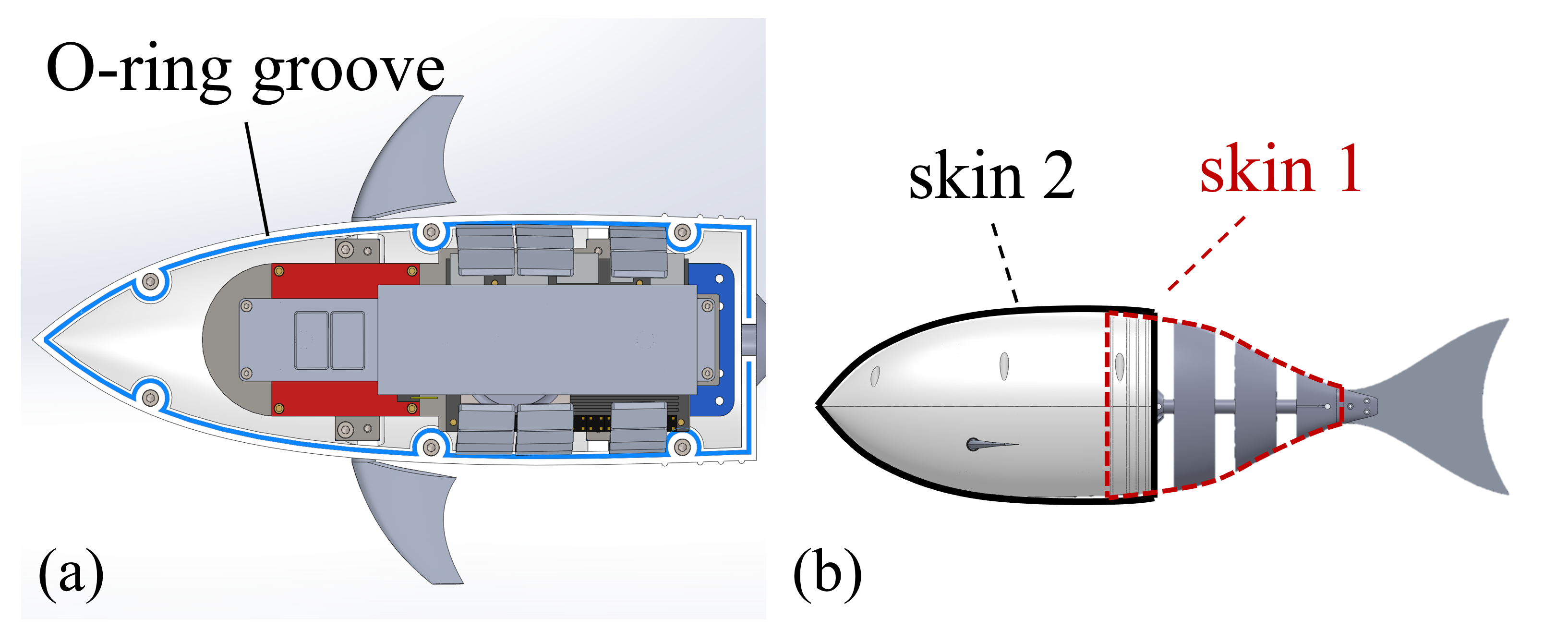}
    \caption{Waterproof design of the fish: (a) O-ring based primary waterproof structure for the rigid head; (b) the skin composed of two sections that enclose the entire body of the robotic fish.}
    \label{fig:waterproof}
\vspace{-5mm}
\end{figure}
An O-ring based primary waterproof structure is designed to protect the electronic units inside the rigid head, as shown in Fig. \ref{fig:waterproof}\textcolor{blue}{a}. Besides, since the actuation mechanism is located inside the head, the transmission wires must pass through the head shell to drive the body and the outlet becomes a leakage point. Thus, we design a waterproof \textit{skin~1} covering the segmented body sections to provide reliable waterproofing and preserve the streamlined profile of the fish body and its frontal area, as shown in Fig. \ref{fig:waterproof}\textcolor{blue}{b}. But it leaves seams in the head exposed to water ingress. To resolve this, we apply an additional waterproof \textit{skin~2} that wraps around the head shell. This design enables operation in shallow water (depths below 1 m) while preserving a streamlined profile and improving internal space utilization, supporting future development.

\section{Modeling and Analysis}\label{sec:modeling}

\subsection{Kinematic Model of the Actuation Mechanism}\label{subsec:kine_act_mech}
The input of the actuation mechanism consists of the angular positions and velocities of the two driving motors. The rotation of the corresponding wheels directly determines the reel motion, which in turn drives the body and overall motion of the robotic fish. As shown in Fig. \ref{fig:kine_act}, two driving wheels of radius $R_a$ are mounted at lateral offsets $L_1$ and $L_2$ relative to the reel of radius $R_b$. Let the angular positions of the two motors be denoted by $\theta_1$ and $\theta_2$, and the angular position of the reel be denoted by $\theta_3$. The positions of the two wheel joints and the reel joint are constrained to lie on the same line of the connecting rod. This collinearity condition can be expressed by the straight-line equation
\begin{equation}
\begin{split}
    &\frac{R_a \sin \theta_2 - R_b \sin \theta_3}{(R_a \cos \theta_2) - (R_b \cos \theta_3 + L_2)}
    \\[5pt]
    =&\frac{R_a \sin \theta_1 - R_b \sin \theta_3}{(R_a \cos \theta_1 + L_1 + L_2 )-(R_b \cos \theta_3 + L_2)}.
\end{split}\label{eq:line}
\end{equation}
Solving \eqref{eq:line} yields a closed-form expression for the reel angle $\theta_3$ in terms of the wheel angles $\theta_1$ and $\theta_2$:
\begin{subequations}
\label{eq:solution}
\begin{align}
\theta_3 &= \arcsin\left( \frac{C}{\sqrt{A^2 + B^2}} \right) - \arctan\left( \frac{B}{A} \right), \\
A &= R_a R_b (\cos \theta_1 - \cos \theta_2)+R_b (L_1+L_2), \\
B &= R_a R_b (\sin \theta_2 - \sin \theta_1), \\
C &= R_a^2 \sin(\theta_1 - \theta_2) - R_a (\sin \theta_1 L_2 + \sin \theta_2 L_1).
\end{align}
\end{subequations}

The relative motion of the two wheel joints and the reel joint can then be determined by rigid-body kinematics. From this relation, the instantaneous angular velocity of the reel, $\dot{\theta}_3$, can be expressed as
\begin{equation}
\label{eq:reel_omega}
    \dot{\theta}_3 = \frac {v_{1n} + \left( v_{1n}-v_{2n}\right) \left(\frac{d_1}{d_1+d_2}\right)} {R_b\cos(\theta_3-\theta_{\text{rod}})},
\end{equation}
where $d_1$ and $d_2$ are the distances from the reel joint to wheel joints 1 and 2, $\theta_{\text{rod}}$ is the orientation of the connecting rod, and $v_{1n}$ and $v_{2n}$ are the velocity components of the wheel joints along the normal to the rod. These quantities are given by
\begin{subequations}
\begin{align}
    \theta_{\text{rod}} &= \arctan\left(\frac{R_a\sin\theta_1-R_a\sin\theta_2}{L_1+L_2+R_a\cos\theta_1-R_a\cos\theta_2}\right), \\
    v_{1n} &= \dot{\theta}_1 R_a\cos(\theta_1 - \theta_{\text{rod}}),\\
    v_{2n} &= \dot{\theta}_2 R_a\cos(\theta_2 - \theta_{\text{rod}}).
\end{align}
\end{subequations}

\begin{figure}[b]
    \centering
    \includegraphics[width=\linewidth]{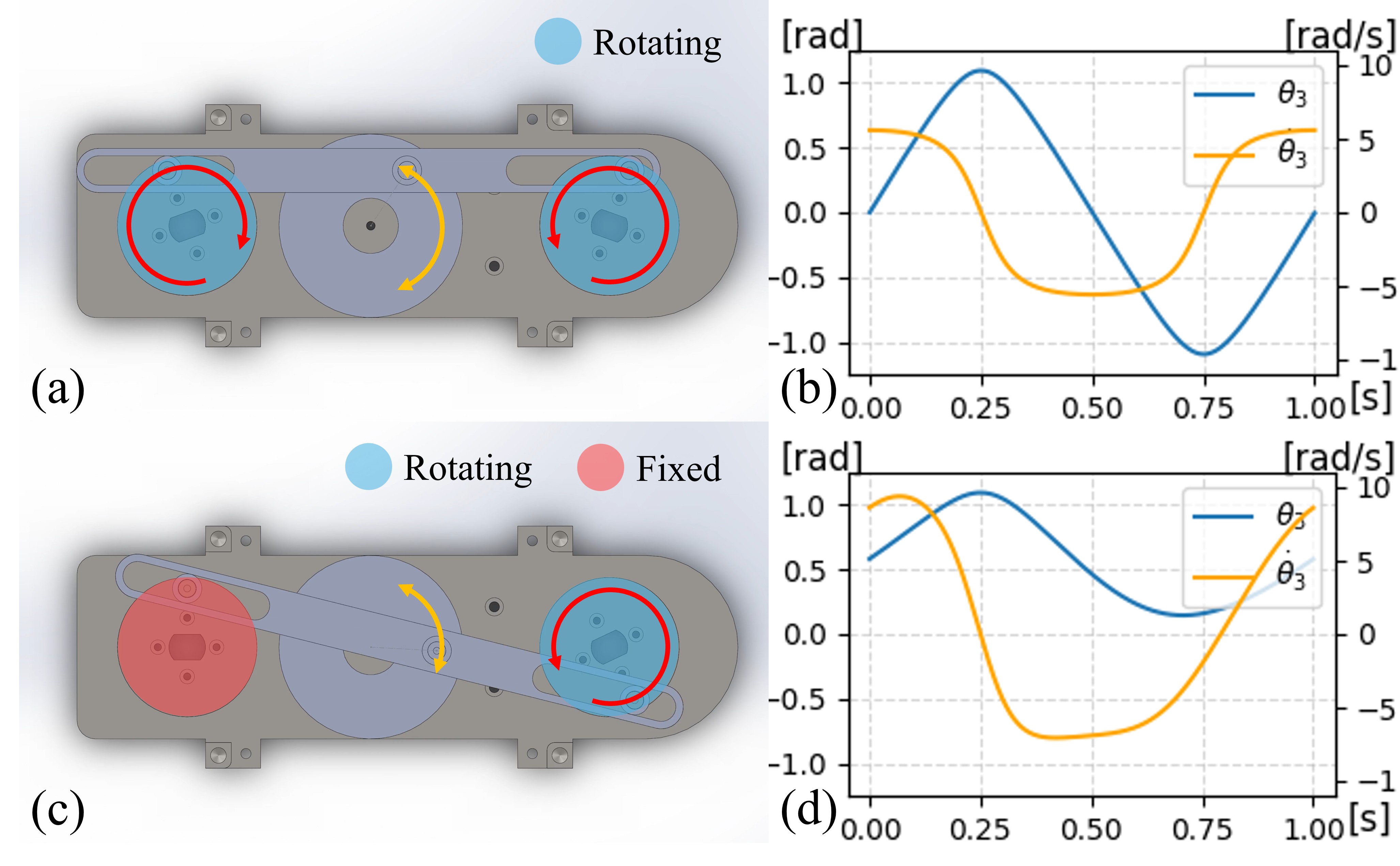}
    \caption{Operating modes of the 2-DoF crank–slider actuation mechanism for directional control. (a) Principle of the symmetric mode; (b) reel output in symmetric mode; (c) principle of the asymmetric mode; (d) reel output in asymmetric mode.}
    \label{fig:motion modes}
\end{figure}

The 2-DoF actuation mechanism has various motion properties. Both of the motors could be worked at position control mode or speed control mode. Thus, in order to make use of this property to emphasize the function of the design, we define two motion modes of the actuation mechanism: the symmetric mode and the asymmetric mode, as shown in Fig. \ref{fig:motion modes}.

In the symmetric mode, as shown in Fig. \ref{fig:motion modes}\textcolor{blue}{a}, both motors rotate with the same angular position and velocity, i.e., $\theta_1 = \theta_2$ and $\dot{\theta}_1 = \dot{\theta}_2=\omega$, where $\omega$ denotes the common motor angular velocity. Under this condition, the entire mechanism degenerates into an ordinary 1-DoF construction, and the reel motion $[\theta_3, \dot{\theta}_3]$ can be expressed as a function of $\omega$:
\begin{equation}\label{eq:f1}
    [\theta_3, \dot{\theta}_3] = f_1(\omega),
\end{equation}
where $f_1$ can be derived from \eqref{eq:solution} and \eqref{eq:reel_omega} with input $\omega$, as shown in Fig. \ref{fig:motion modes}\textcolor{blue}{b}.

In the asymmetric mode, Fig. \ref{fig:motion modes}\textcolor{blue}{a}, motor 1, the red, rotates continuously while motor 2, the blue, remains fixed at a certain angular position. In this case, $\dot{\theta}_1=\omega$ and $\theta_2 = \text{constant}$, and the reel motion becomes a function of both $\theta_2$ and $\omega$:
\begin{equation}\label{eq:f2}
    [\theta_3, \dot{\theta}_3] = f_2(\theta_2,\omega),
\end{equation}
where $f_2$ can be derived from \eqref{eq:solution} and \eqref{eq:reel_omega} with input $\theta_2$ and $\omega$, as shown in Fig. \ref{fig:motion modes}\textcolor{blue}{d}.

By combining the kinematic relation \eqref{eq:solution} and the velocity equation \eqref{eq:reel_omega}, the tendon displacement and velocity can be explicitly derived from the motor angles and angular velocities. These quantities provide the actuation inputs to the elastic body model, forming the basis for subsequent dynamic analysis.

\subsection{Dynamic Model of the Elastic Body}\label{subsection: modeling}
\begin{figure}[b]
    \centering
    \includegraphics[width=0.7\linewidth]{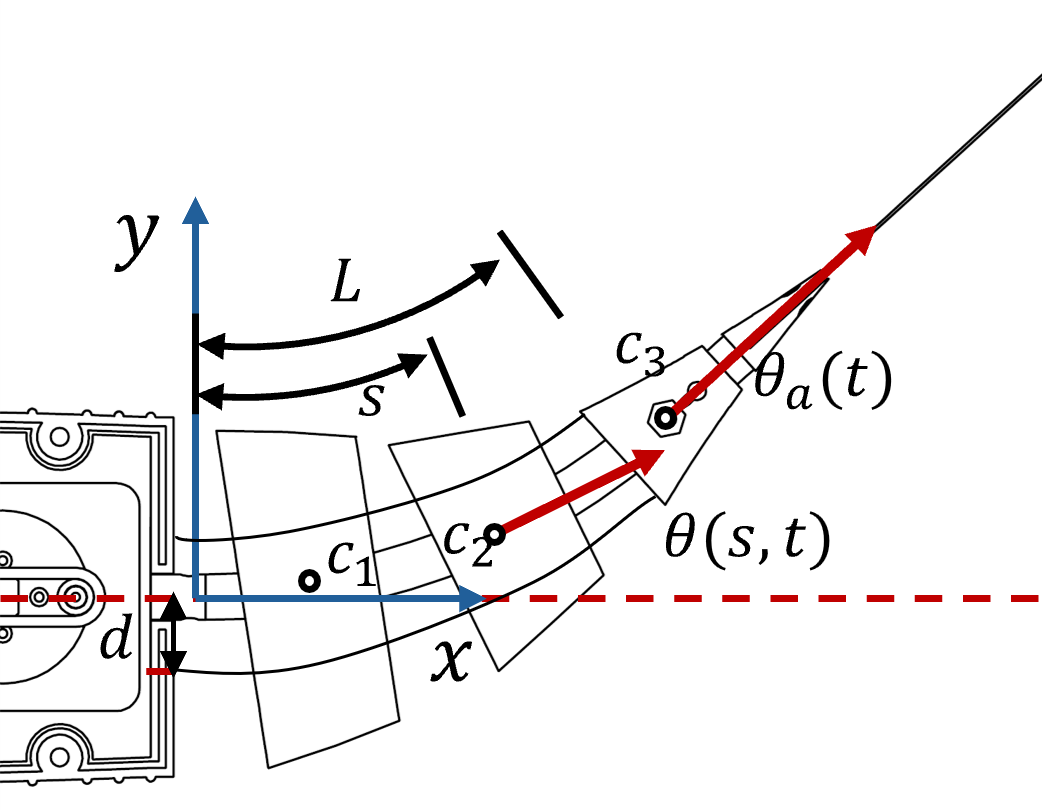}
    \caption{Dynamic model of the elastic tail.}
    \label{fig:dyna_tail}
\vspace{-5mm}
\end{figure}

To estimate the motor load required for different oscillation states and provide a criterion for motor selection, we construct a dynamic model of the elastic body. The analysis focuses only on the body sections while neglecting head rotation, which simplifies the formulation while preserving the essential characteristics of caudal propulsion, as illustrated in Fig. \ref{fig:dyna_tail}.

The elastic body is actuated by a pair of wires attached in parallel to the spine at its trailing end. These wires generate a pure torque about the end of the spine, with an offset distance $d$ from the centerline of the spine. Therefore, an important assumption of the model is that the spine bends into an arc of constant curvature and that the driving wire remains parallel to the spine. Then, the angle of attack $\theta_a$ and its angular velocity $\dot{\theta}_a$ can be related to the reel motion from \eqref{eq:solution} and \eqref{eq:reel_omega} as
\begin{equation}\label{eq:theta_a}
    \theta_a(t) = \frac{R_b \theta_3(t)}{d},
\end{equation}
\vspace{-5mm}
\begin{equation}\label{eq:dtehta_a}
    \dot{\theta}_a(t) = \frac{R_b \dot{\theta}_3(t)}{d}.
\end{equation}
\vspace{-5mm}

For a rod bending with constant curvature, the angle $\theta(s,t)$ and position $\bm{r}(s,t)$ of a point located at arc length $s$ from the fixed end can be expressed in terms of the attack angle $\theta_a$:
\begin{equation}
    \theta(s,t) = \frac{\theta_a(t)}{L}s,
\end{equation}
\begin{equation}
    \bm{r}(s,t) = \left[ \frac{L}{\theta_a}\sin\left(\theta(s,t)\right), \frac{L}{\theta_a}\left(1-\cos\left(\theta(s,t)\right)\right)\right],
\end{equation}

Thus, the speed and acceleration of the point could be represented as 
\begin{equation}
    \dot{\bm{r}}(s,t) = \frac{\partial\bm{r}}{\partial\theta_a}\dot{\theta}_a,
\end{equation}
\begin{equation}
    \ddot{\bm{r}}(s,t) = \frac{\partial\bm{r}}{\partial\theta_a}\ddot{\theta}_a + \frac{\partial^2\bm{r}}{\partial\theta_a^2}\dot{\theta}_a^2.
\end{equation}

The motors are controlled through their rotational speeds; therefore, the dynamic model is primarily established to evaluate the motor load under different flapping frequencies. Because the elastic spine does not bend uniformly along its length, an equivalent average spring rate $K_{\text{eq}}$ is introduced. The resulting elastic moment is modeled as
\begin{equation}
M_{\text{elastic}} = K_{\text{eq}}\theta_a ,
\end{equation}
each section of the tail has weight $m_i$ and the center of mass (CoM) is located at arc length $c_i$ along the rod. The added mass can be estimated from the density of water $\rho$, the reference diameter of the body sections $D(s)$ with respect to the arc length $s$ and the Jacobian  $J(s,\theta_a)$. Therefore, the total kinetic energy of the tail can then be written as
\begin{subequations}
\begin{align}
T &= \frac{1}{2} \dot{\theta}_a^2 \left[ \int_0^L\rho D(s) \|J(s,\theta_a)\|^2\mathrm{ds} + \sum_i m_i\|J_i\|^2\right], \\
& J(s,\theta_a) = \frac{\partial\bm{r}}{\partial\theta_a}, \\
& J_i = J(c_i,\theta_a).
\end{align}
\end{subequations}
The potential energy stored in the elastic spine is given by
\begin{equation}
    U = \frac{1}{2}K_{\text{eq}}\theta_a^2.
\end{equation}

According to Lagrangian mechanics, the equation of motion is given by
\begin{equation}\label{eq:Lag}
    \frac{\mathrm{d}}{\mathrm{d}t}\left(\frac{\partial T}{\partial \dot{\theta_a}}\right)-\frac{\partial T}{\partial\theta_a} + \frac{\partial U}{\partial \theta_a}=Q_{\text{nc}},
\end{equation}
where $Q_{\text{nc}}$ represents the generalized non-conservative torque, including fluid drag $Q_{\text{drag}}$, structural damping $Q_{\text{damp}}$, and the actuation moment $M_{\text{wire}}$. The generalized fluid drag torque is expressed as
\begin{align}
Q_{\text{drag}} &= -\frac{1}{2}\rho C_d 
\int_0^L D(s)\,\dot{\theta}_a \left|\dot{\theta}_a\right| 
\left\|J(s,\theta_a)\right\|^3 \, \mathrm{ds} \notag \\[6pt]
&\quad -\frac{1}{2}\rho C_d A_{\text{fin}}\,\dot{\theta}_a 
\left|\dot{\theta}_a\right|\left\|J(L,\theta_a)\right\|^3
\end{align}
where $C_d$ is the drag coefficient and $A_{\text{fin}}$ is the reference area of the caudal fin. The structural damping torque is modeled as
\begin{equation}
Q_{\text{damp}} = -C_{\text{damp}}\dot{\theta}_a,
\end{equation}
with $C_{\text{damp}}$ denoting the damping coefficient. Finally, the actuation moment generated by the driven wire can be calculated from a second-order nonlinear ODE as
\begin{equation}
\label{eq:final}
    M_{\text{wire}}= \frac{\mathrm{d}}{\mathrm{d}t}\left(\frac{\partial T}{\partial \dot{\theta_a}}\right)-\frac{\partial T}{\partial\theta_a} + \frac{\partial U}{\partial \theta_a} - Q_{\text{drag}} -Q_{\text{damp}}.
\end{equation}

By solving \eqref{eq:final}, the tendon tension $T_{\text{wire}}$ can be solved with 
\begin{equation}
    T_{wire}=\frac{M_{wire}}{d},
\end{equation}
which can be estimated as a function of $\theta_a$, and thus further expressed in terms of the reel motion $\theta_3$. Since the tendon is actuated through the reel by two motors, the motor torques $\tau_1$ and $\tau_2$ can be obtained through the force transmission relations of the crank–slider mechanism:
\begin{equation}
\begin{cases}
    \dfrac{R_a\tau_1}{\cos{(\theta_1-\theta_{\text{rod}})}} + \dfrac{R_a\tau_2}{\cos{(\theta_2-\theta_{\text{rod}})}} = \dfrac{T_{\text{wire}}}{\cos{(\theta_3-\theta_{\text{rod}})}} \\[10pt]
    \dfrac{d_1R_a\tau_1}{\cos{(\theta_1-\theta_{\text{rod}})}} =  \dfrac{d_2R_a\tau_2}{\cos{(\theta_2-\theta_{\text{rod}})}}
\end{cases}\label{eq:motor_load}
\end{equation}

\begin{figure}[b]
    \centering
    \includegraphics[width=1.0\linewidth]{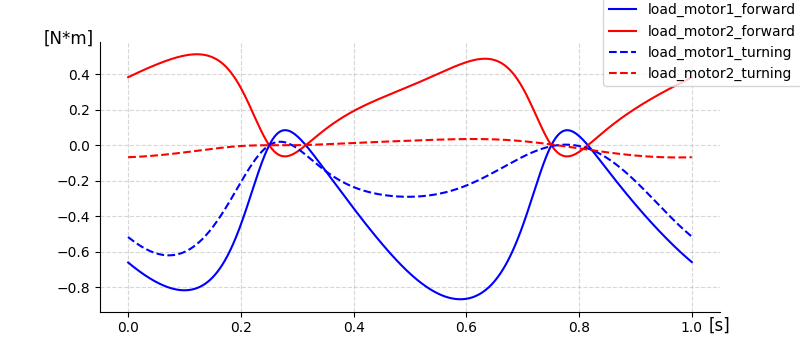}
    \caption{Required motor torque for achieving \SI{1}{Hz} oscillation, estimated using the dynamic model.}
    \label{fig:dyna_plot}
\vspace{-5mm}
\end{figure}

The torque required to achieve \SI{1}{\hertz} oscillation in symmetric mode and asymmetric mode where $\theta_2=\SI{90}{\degree}$ is calculated using \eqref{eq:theta_a}, \eqref{eq:dtehta_a}, \eqref{eq:final}, and \eqref{eq:motor_load}. The result is shown in Fig. \ref{fig:dyna_plot}, assuming a drag coefficient of $C_d=1.2$, rod damping $C_{\text{damp}}=\SI{0.02}{\kilo\gram\metre\squared\per\second\per\radian}$, rod stiffness $K_{\text{eq}}=\SI{0.248}{\newton\metre\per\radian}$, and neglecting friction. This estimation guides motor selection in Section \ref{subsec:proto}.

\section{Control system} \label{sec:control}

\subsection{Feedforward Control} \label{subsec:feedforward}

As presented in Section \ref{subsec:kine_act_mech}, we define two motion modes for the 2-DoF crank-slider mechanism: symmetric mode \eqref{eq:f1} and asymmetric mode \eqref{eq:f2}. 

The symmetric mode has only one input, the rotation speed $\omega$ of both motors. The synchronized rotation of the two motors has the maximum power input and oscillation amplitude, which generates higher thrust force and thus higher swimming speed. We define $g_1$ as the relationship between the swimming speed of the robot $v_{\text{robot}}$ and $\omega$, which could be estimated from \eqref{eq:f1}, \eqref{eq:theta_a}, \eqref{eq:dtehta_a}, as follows:
\begin{equation}\label{eq:g1}
    v_{\text{robot}} = g_1(\theta_3, \dot{\theta}_3)=g_1(f_1(\omega)),
\end{equation}

The asymmetric mode has two inputs, the position of motor 2, $\theta_2$ and the angular speed of motor 1, $\omega$. The displacement of motor 2 forms a non-zero mean angle of oscillation, while the continuous rotation of motor 1 provides continuous input power for oscillation and controls its frequency, which decouples the control of swimming speed $v_{\text{robot}}$ and yaw direction $\psi_{\text{robot}}$. According to \eqref{eq:f2}, \eqref{eq:theta_a}, \eqref{eq:dtehta_a}, the motion of the robot follows:

\begin{equation}\label{eq:g2}
    [v_{\text{robot}}, \psi_{\text{robot}}] = g_2(\theta_3, \dot{\theta}_3)=g_2(f_2(\theta_2,\omega)).
\end{equation}

The asymmetric mode allows the fish robot to control the yaw direction and oscillation frequency with different actuators, which decouples the propulsion and steering of swimming and enables smooth turning. 

\subsection{Feedback Direction Control} \label{subsec:feedback}
\begin{figure}[t]
    \centering
    \includegraphics[width=\linewidth]{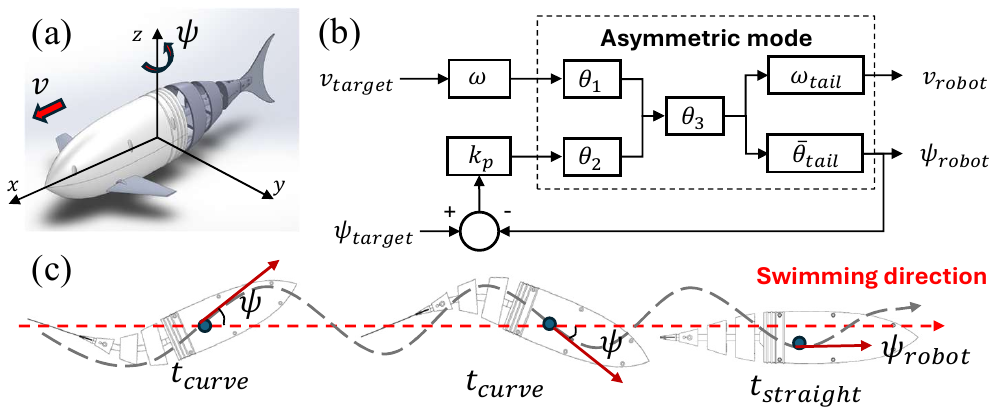}
    \caption{Control framework of the robotic fish. (a) Controlled volume and reference frames; (b) block diagram of the control system; (c) swimming direction during body oscillation.}
    \label{fig:feedback control}
\vspace{-5mm}
\end{figure}

Direction control is a fundamental function for further precise maneuvering, which allows the robot to overcome model error and disturbance from the surroundings. To address this, the present study exploits the asymmetry of the driving mechanism to achieve basic yaw-directional control, as shown in Fig. \ref{fig:feedback control}\textcolor{blue}{a}. As shown in Fig. \ref{fig:feedback control}\textcolor{blue}{b}, the position of the fixed wheel in the asymmetric mode changes according to the real angle $\psi_{\text{robot}}$ and the reference angle $\psi_{\text{target}}$, which can be expressed as
\begin{equation}
    \theta_2 = k_p(\psi_{\text{target}}-\psi_{\text{robot}}),
\end{equation}
while the rotating wheel controls the oscillation frequency $\omega$ to control the swimming speed $v_{\text{robot}}$ can be estimated by \eqref{eq:g2}.

The key problem in direction control is the definition of the swimming direction $\psi_{\text{robot}}$. In the process of undulatory propulsion, the head of the robotic fish inevitably oscillates along with the body, and the direction measured $\psi(t)$ changes. We observed that during periodic oscillations, when the tail deflection passes through its mean position $t_{\text{mean}}$ and continuous rotating motor 1 passes the mean position $\theta_1=0$, the orientation of the head is approximately aligned with the tangent direction of the swimming trajectory, as shown in Fig. \ref{fig:feedback control}\textcolor{blue}{c} can be expressed as
\begin{equation}
    \psi_{\text{robot}} := \psi(t_{\text{mean}})
\end{equation}
Based on this observation, we define the measured heading angle at the instant when the rotating wheel returns to its neutral position and update $k_p$ at this state.

\section{Experiment} \label{sec:exp}
\subsection{Prototype and Equipment Specification}\label{subsec:proto}

\begin{figure}[t]
    \centering
    \includegraphics[width=\linewidth]{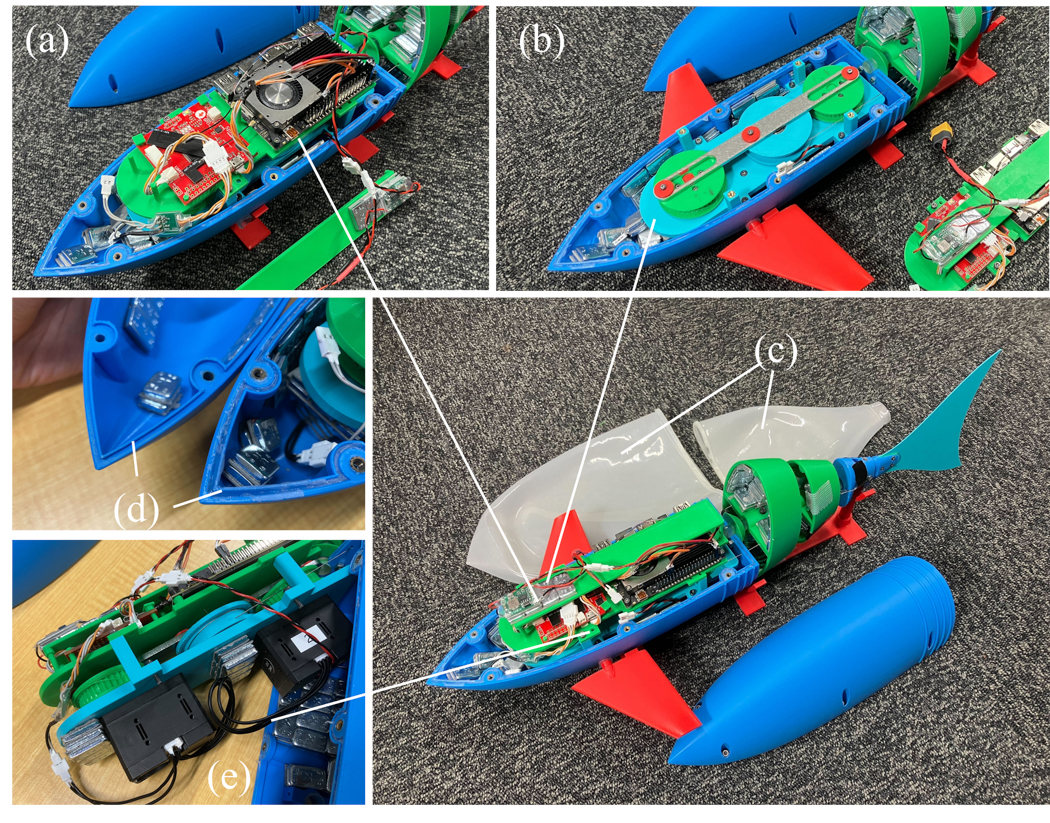}
    \caption{Prototype of the robotic fish and its key components. (a) microcontroller unit (MCU) and onboard PC; (b) actuation mechanism; (c) waterproof silicone skin; (d) waterproof structure of the rigid head shell; (e) motors used in the prototype.}
    \label{fig:prototype}
\vspace{-3mm}
\end{figure}

We developed a prototype of the robotic fish, as shown in Fig.~\ref{fig:prototype}. The prototype measures \SI{502}{\milli\meter} in length, \SI{83}{\milli\meter} in width, \SI{128}{\milli\meter} in height, and has a mass of \SI{2050}{\gram}. The rigid head, body sections, and caudal fin are fabricated from PLA using 3D printing, while the actuation linkage is made of SUS304 stainless steel. The elastic spine, made of urethane, has a diameter of \SI{10}{\milli\meter} and a Shore hardness of \SI{70}{\ampere}. For waterproofing, silicone is incorporated into the sealing design: a groove–protrusion interface along the head shell holds a silicone layer to protect the electronics, and a silicone skin covers the entire body to prevent leakage from the wire outlet.

The actuation system consists of two servo motors (DYNAMIXEL XC340-T150-BB-T) that drive the wheels in continuous rotation mode. Each motor delivers \SI{1.6}{\newton\meter} torque—twice the estimated requirement in Section~\ref{subsection: modeling}—and achieves a tail-beat frequency of \SI{1}{\hertz}. Power is supplied by a 3S \SI{1800}{\milli\ampere\hour} LiPo battery, enabling approximately \SI{50}{\minute} of swimming.

\begin{figure}[t]
    \centering
    \includegraphics[width=0.9\linewidth]{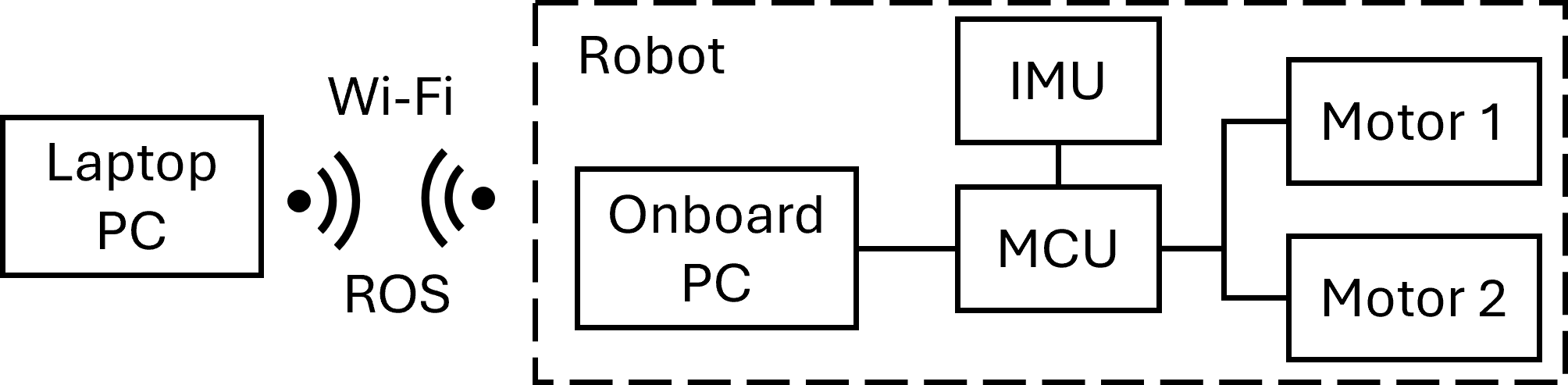}
    \caption{Software architecture of the robotic fish.}
    \label{fig:software}
\vspace{-5mm}
\end{figure}

The control and communication system is shown in Fig.~\ref{fig:software}. A laptop PC issues commands to an onboard PC (Khadas VIM4), which works together with a custom MCU integrates an STM32H7 chip with IMU and geomagnetic sensors for direction estimation. Communication between the laptop and onboard PC is established via \SI{2.4}{\giga\hertz} WiFi using the ROS framework. This system supports operation near the water surface.

\subsection{Oscillation in Air}
\begin{figure}[b]
    \centering
    \includegraphics[width=\linewidth]{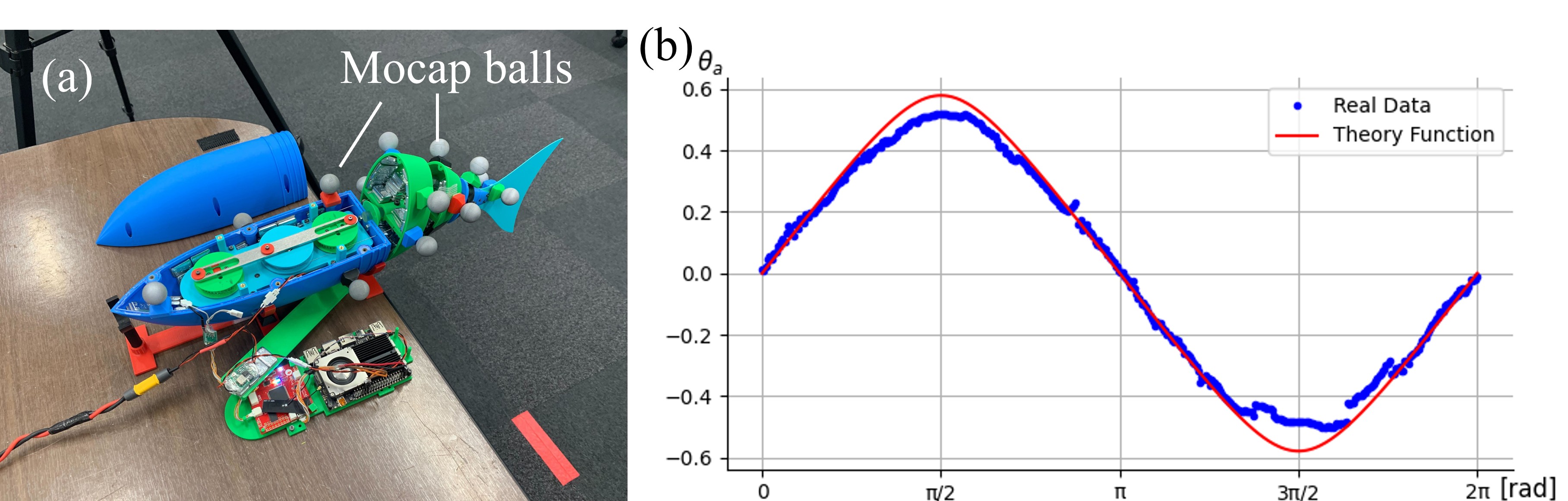}
    \caption{Flapping-in-air experiment for validating the theoretical flapping angle of the robotic fish. (a) Motion capture balls attached to each body section to measure position and orientation; (b) comparison between theoretical prediction and experimental data.}
    \label{fig:in_air}
\end{figure}
To validate the dynamic model presented in Section \ref{subsection: modeling}, we conducted an in-air test using a motion capture system, as shown in Fig. \ref{fig:in_air}. Reflective markers were attached to the robot body for camera tracking. The theoretical displacement obtained from \eqref{eq:solution} is compared with the measured values, which confirms the accuracy of the geometric model of the elastic body. Based on the motor load data, the average equivalent spring rate was calculated as $K_{\text{eq}} = 0.248\ \text{N·m}/\text{rad}$, and the load response under motion was also evaluated.

\subsection{Swimming Performance} \label{subsec:exp_swim}
\begin{figure}
    \centering
    \includegraphics[width=0.87\linewidth]{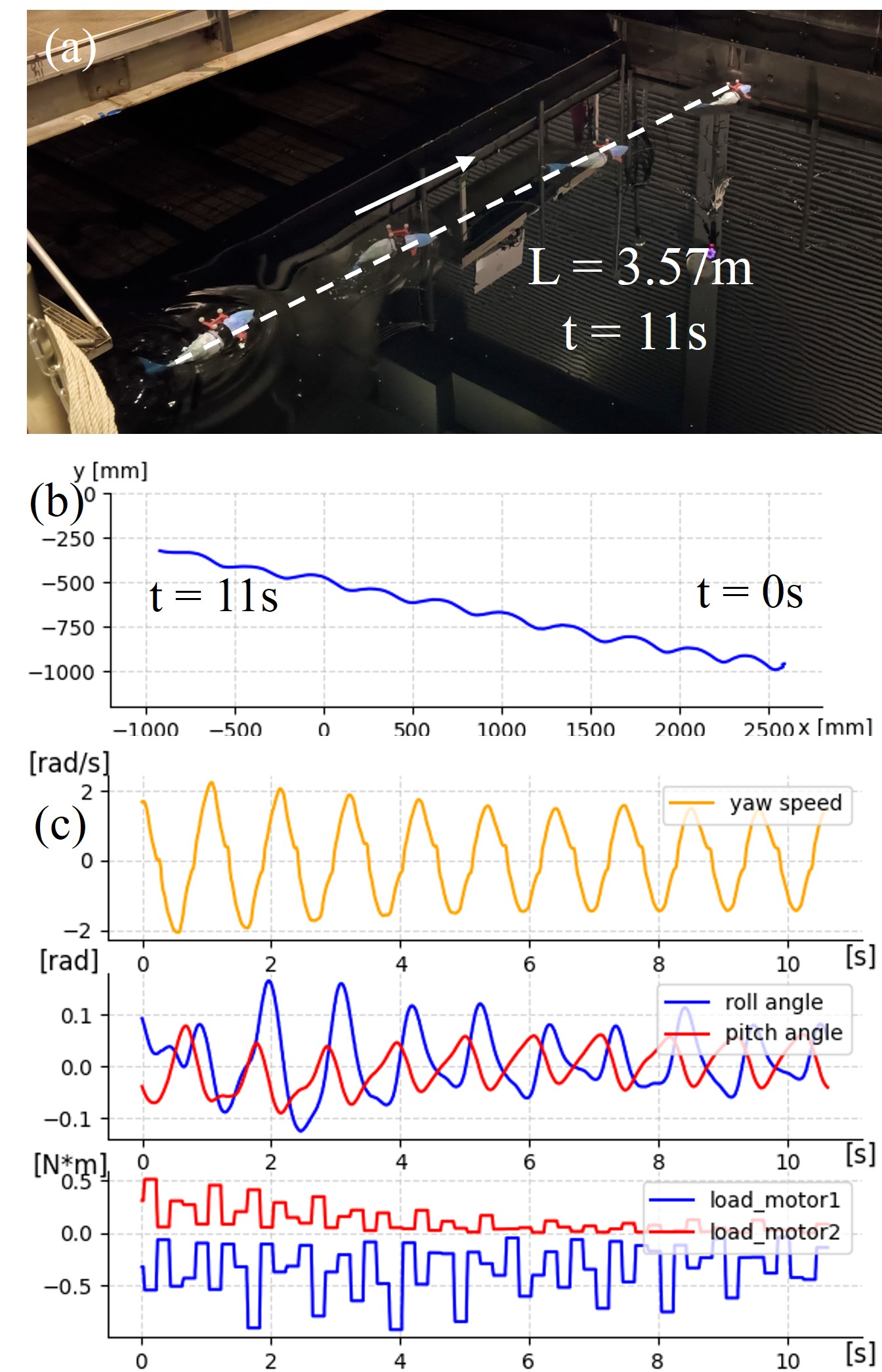}
    \caption{Forward swimming experiment of the robotic fish in symmetric mode. (a) Swimming trajectory recorded in a water tank; (b) real-time position obtained from motion capture; (c) IMU measurements and motor loads during swimming.}
    \label{fig:exp_swim}
\vspace{-5mm}
\end{figure}

To verify the motion capability of the prototype, we conducted experiments in a pool measuring \SI{5}{\meter} by \SI{4}{\meter}, where a motion capture system recorded the real-time position of the robot. From these measurements, we calculated the maximum forward speed and turning speed of the prototype.

\subsubsection{Forward Swimming Test}
The first experiment examined forward swimming in the symmetric mode presented as \eqref{eq:g1}, where both motors rotated simultaneously at \SI{1}{\hertz}. Under this condition, the robot traveled \SI{3.75}{\meter} in \SI{11}{\second}, achieving an average speed of \SI{0.32}{\meter\per\second} (\SI{0.64}{BL/s}), as shown in Fig.~\ref{fig:exp_swim}\textcolor{blue}{a}. To further evaluate the propulsion efficiency, we measured that the peak-to-peak tail amplitude is \SI{0.25}{\meter}, yielding a Strouhal number of $St=0.78$. The relatively high Strouhal number reflects the current operating condition of the prototype and indicates potential room for efficiency optimization.

\subsubsection{Turning Test}
The second experiment evaluated turning performance in the asymmetric mode as \eqref{eq:g2}, where wheel 2 was fixed at $-90^\circ$ and wheel 1 rotated at \SI{1}{\hertz}. In this configuration, the robot achieved a turning radius of \SI{0.56}{\meter} (\SI{1.12}{BL}), completing one full circle in \SI{14}{\second} with an average turning speed of \SI{25.7}{\degree\per\second}, as shown in Fig.~\ref{fig:exp_turn}\textcolor{blue}{a}.

\subsubsection{Motion Stability}
The motion states during forward swimming and continuous turning are illustrated in Figs.~\ref{fig:exp_swim}\textcolor{blue}{c} and \ref{fig:exp_turn}\textcolor{blue}{c}. In both cases, the maximum roll and pitch angles remained below \SI{0.16}{\radian}, while the oscillatory yaw rate was about \SI{2.24}{\radian\per\second} in symmetric mode and \SI{1.65}{\radian\per\second} during turning. These results demonstrate stable swimming and smooth maneuvering, confirming the feasibility of the hardware design and the effectiveness of the decoupled propulsion and steering control.

\subsubsection{Motor Load Analysis}
\begin{figure}[t]
    \centering
    \includegraphics[width=0.87\linewidth]{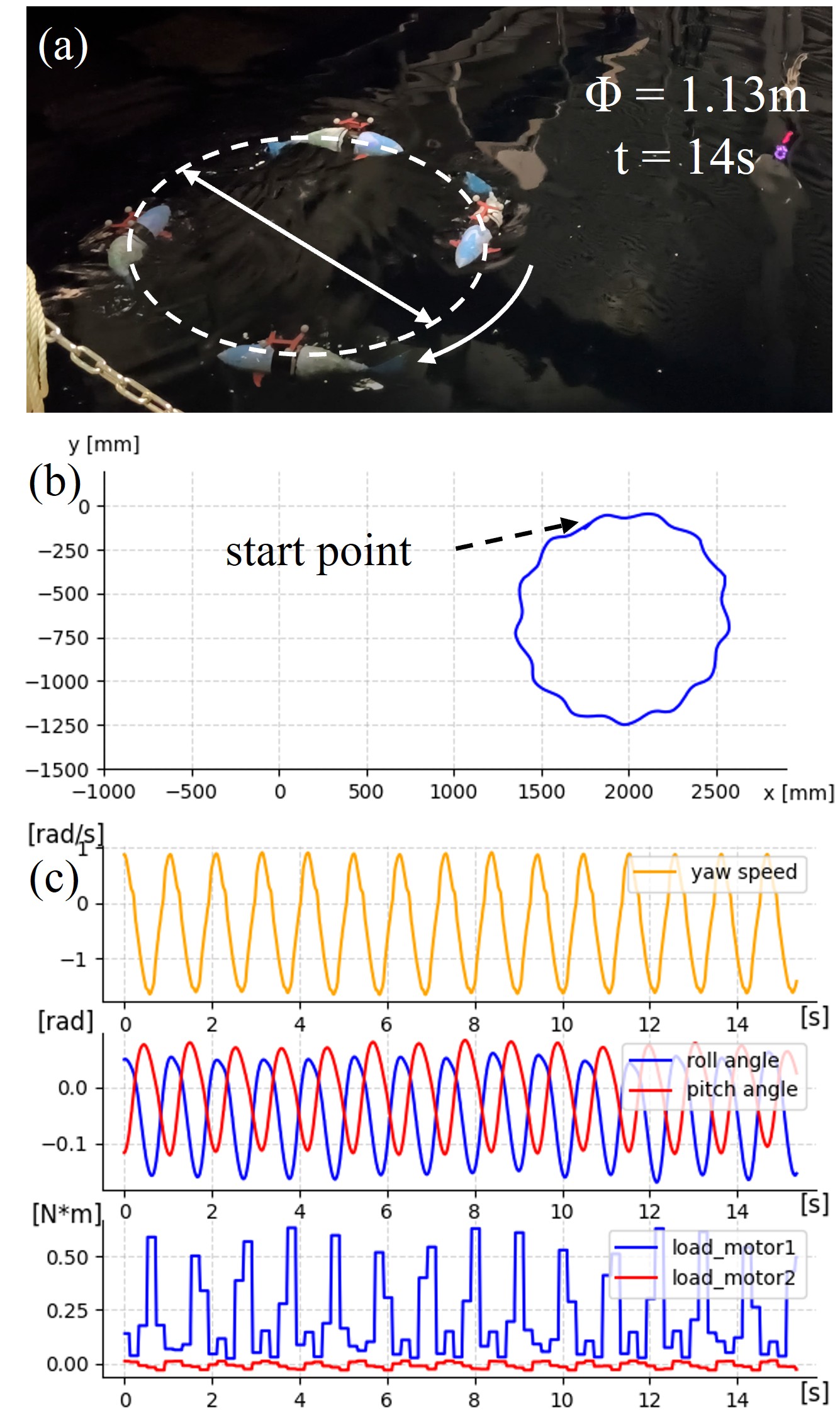}
    \caption{Turning experiment of the robotic fish in asymmetric mode. (a) Circular swimming trajectory in a water tank; (b) real-time position recorded by motion capture; (c) IMU measurements and motor loads during turning.}
    \label{fig:exp_turn}
\vspace{-5mm}
\end{figure}

Figs.~\ref{fig:exp_swim}\textcolor{blue}{c} and \ref{fig:exp_turn}\textcolor{blue}{c} also show the motor loads. During forward swimming and continuous turning, the maximum torques of motor 1 and motor 2 were \SI{0.92}{\newton\meter} and \SI{0.51}{\newton\meter}, respectively. Although the measured values exceeded model predictions due to empirical drag coefficients and neglected friction, the consistency of the trends confirms that the dynamic model remains effective for estimating motion states and provided a feasible criterion for selecting the hardware, such as the spine and motors.

\subsection{Feedback Control Performance}
We implemented feedback-based direction control using the IMU and magnetometer, as introduced in Section~\ref{subsec:feedback}. We applied the asymmetric mode as introduced in Section~\ref{subsec:feedforward} and Section~\ref{subsec:exp_swim} that decouples the propulsion and enabled the smooth turning during continuous propulsion.

\subsubsection{Trajectory Tracking}
\begin{figure}[t]
    \centering
    \includegraphics[width=0.87\linewidth]{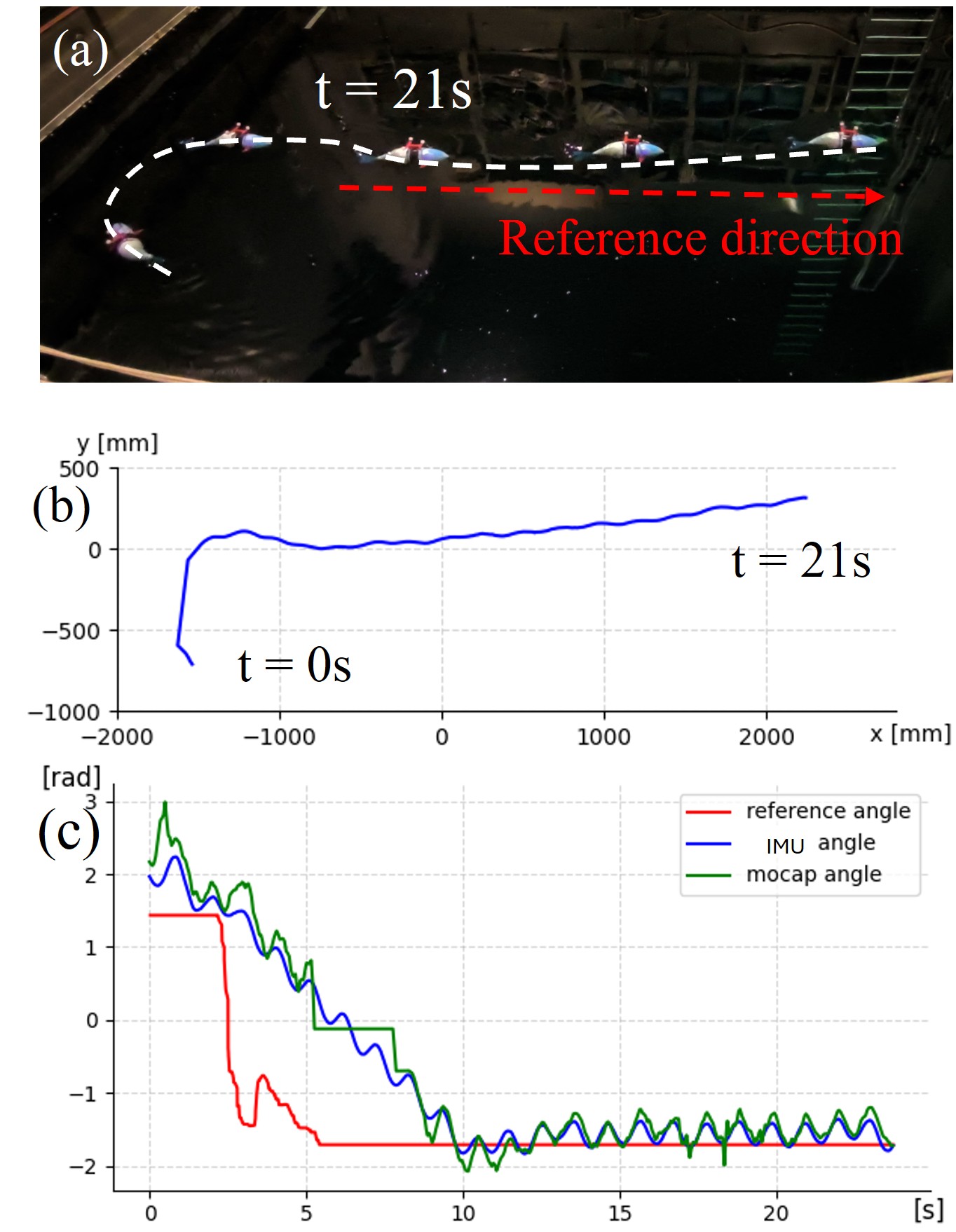}
    \caption{Direction control experiment of the robotic fish in asymmetric mode. (a) Swimming trajectory in a water tank; (b) real-time position recorded by motion capture; (c) comparison of reference, actual, and motion-capture angles.}
    \label{fig:exp_imu}
\vspace{-5mm}
\end{figure}

The motion snapshots and recorded trajectory from the motion capture system are shown in Fig.~\ref{fig:exp_imu}\textcolor{blue}{a} and Fig.~\ref{fig:exp_imu}\textcolor{blue}{b}. Fig.~\ref{fig:exp_imu}\textcolor{blue}{c} compares the direction estimated by \textit{Spinal} with the reference angle provided by the operator, confirming the feasibility of the actuation mechanism and the effectiveness of the decoupled control strategy.

\subsubsection{Control Accuracy and Response}
The mean angle measured by the MPU was \SI{-1.56}{\radian}, while the reference angle was \SI{-1.71}{\radian}, resulting in a steady-state error of \SI{0.15}{\radian}. This error was small and convergent, demonstrating the validity of the direction control. To further improve accuracy, an integral term could be introduced to eliminate the residual error. The system exhibited a response time of \SI{8.3}{\second} and a rotation speed of \SI{21}{\degree\per\second}, which enabled a U-turn within \SI{8.3}{\second}.

This feedback control experiment verifies that the asymmetric mode, together with the proposed actuation mechanism, enables effective regulation of swimming direction and speed through decoupled control of oscillation angle and frequency.

\section{Conclusion} \label{sec:conclusion}
This work presents a wire-driven robotic fish featuring a novel 2-DoF crank–slider actuation mechanism that decouples propulsion and steering, supported by a structural and waterproof design. We model the actuation mechanism and elastic body with fluid dynamics, enabling estimation of motor loads under different locomotion states and guiding motor selection. Based on this foundation, we develop a control system tailored to the motion characteristics of the mechanism.

Experiments on swimming, turning, and feedback-based direction control validate the proposed design and demonstrate effective decoupled control of mean oscillation angle and frequency, allowing independent regulation of direction and speed.

A current limitation is that high-frequency performance (up to \SI{6}{\hertz}) remains untested, which can be addressed with more powerful brushless motors. Future work will also explore richer motion modes and more advanced control strategies beyond proportional control to further enhance maneuverability.



\footnotesize
\bibliographystyle{bibtex/IEEEtran}
\bibliography{bibtex/IEEEabrv, bibtex/references}

\normalsize
\end{document}